\documentclass{article}
\usepackage{iclr2027_conference,times}

\usepackage{amsmath,amsfonts,bm}

\def\eqref#1{equation~\ref{#1}}

\def\1{\bm{1}}

\DeclareMathAlphabet{\mathsfit}{\encodingdefault}{\sfdefault}{m}{sl}
\SetMathAlphabet{\mathsfit}{bold}{\encodingdefault}{\sfdefault}{bx}{n}

\usepackage{hyperref}
\usepackage{url}
\usepackage{booktabs}
\usepackage{amsfonts}
\usepackage{nicefrac}
\usepackage{microtype}
\usepackage{xcolor}
\usepackage{kotex}
\usepackage{amsmath}
\usepackage{graphicx}
\usepackage{float}
\usepackage{pgfplots}
\pgfplotsset{compat=1.18}
\usepackage{tabularx}
\usepackage{array}

\definecolor{oursgreen}{HTML}{4F9C68}

\title{Recursive Harness Distillation across Agents for Robot Manipulation}

\author{
  Seungyeon Kim$^{1}$ \quad Junhoo Lee$^{2}$ \quad Minkyu Kim$^{1}$ \quad Baekseung Kim$^{1}$ \quad Nojun Kwak$^{1}$ \\[4pt]
  $^{1}$Seoul National University \\
  $^{2}$Korea Advanced Institute of Science and Technology \\[4pt]
  \texttt{\{syeonkim,minkyu.kim,baekseung.kim\}@snu.ac.kr} \\
  \texttt{junhoo.lee@kaist.ac.kr}
}

\iclrfinalcopy

\begin{document}

\maketitle
\lhead{}
\begin{abstract}
A central goal in robotics is to enable manipulation across changing tasks and environments. Vision-language-action (VLA) models provide broad manipulation capabilities but can struggle when execution requires diagnosing failures and adapting behavior. Strong agents can discover effective interventions through interaction with these policies. We propose \textbf{Recursive Harness Distillation} to accumulate this experience as reusable guidance across agents. A strong agent distills its experience into a playbook for a light agent, then recursively refines the playbook using the light agent’s execution feedback. The resulting playbook enables agents to reuse accumulated intervention knowledge in new task instances without updating model parameters. In real-world manipulation, the harness improves success from 37.3\% to 64.0\%. On SimplerEnv Bridge, the light agent with the playbook achieves 66.7\% success, compared with 41.7\% for the GR00T-only baseline, and outperforms the strong agent without a playbook. The same playbook also benefits the strong agent, which reaches 79.2\% success. These results demonstrate the feasibility of harness distillation for robotics: intervention experience can be accumulated, refined through execution, and reused across agents to improve manipulation.
\end{abstract}

\section{Introduction}
A persistent vision in robotics is to develop systems that can apply their manipulation capabilities across diverse tasks and environments. Vision-language-action (VLA) models bring this goal closer by combining pretrained visual and linguistic representations with robot action learning~\citep{brohan2023rt,kim2024openvla,black2024pi_0}. Using these policies across changing scenes and extended tasks also requires interpreting goals, monitoring execution, and responding when behavior departs from the intended outcome~\citep{huang2022inner,hu2026matters}. Existing approaches use language models to generate executable robot programs~\citep{liang2023code} and to coordinate learned policies and manipulation tools through planning, monitoring, and recovery~\citep{chen2026volo}.

Understanding the task does not by itself tell an agent how to elicit the intended behavior from a deployed VLA~\citep{jeong2026learning}. Even when the agent identifies the correct subgoal, it must determine how the policy will respond to the scene and how to intervene when execution goes wrong. This requires connecting physical observations, policy behavior, and the effects of interventions. Experience with the policy helps the agent learn which adjustments work in which situations.

Highly capable models such as Astra can use this understanding to interpret scenes, diagnose failures, and adjust execution. However, relying on these models throughout execution is costly, motivating the use of lighter, lower-cost models for deployment. Transferring expertise to these models also consumes resources: the capable model must explore the task, analyze its experience, and prepare useful guidance. This raises a question: under a limited token budget, how can we distill this expertise to help a lighter model effectively operate the VLA?

\begin{figure}[t]
  \centering
  \includegraphics[width=\linewidth]{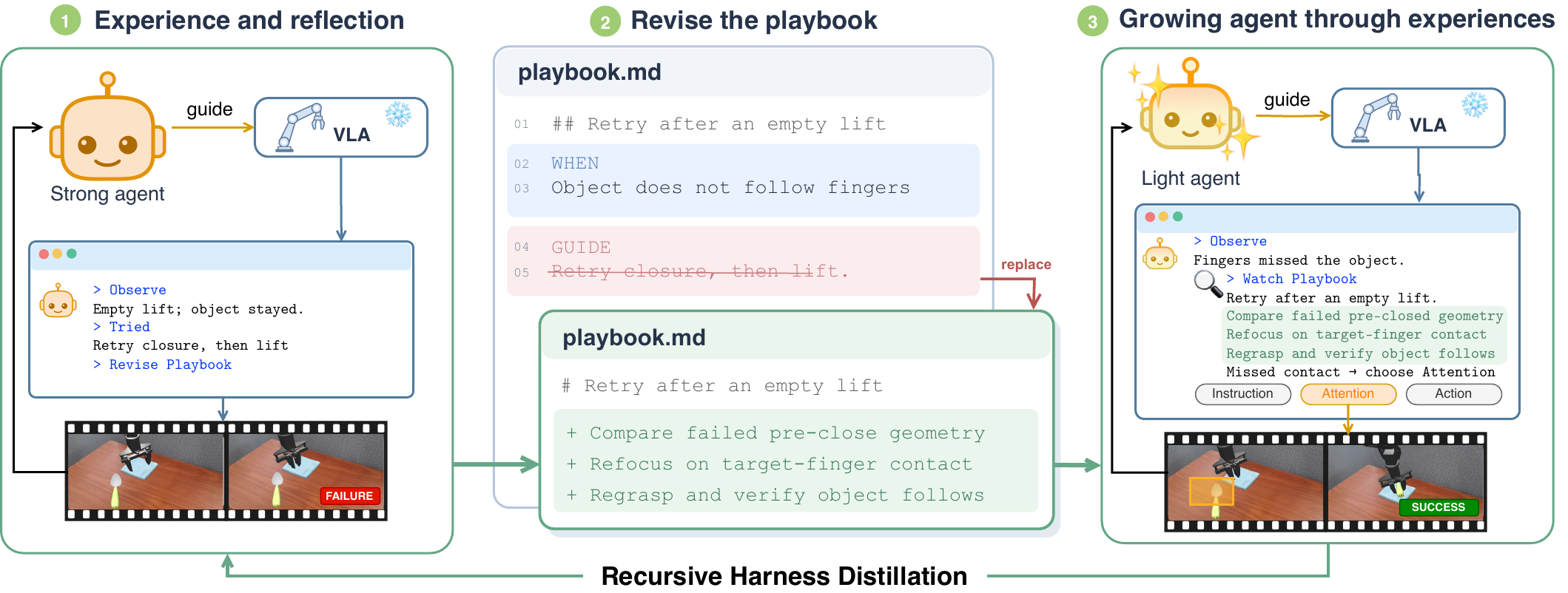}
  \vspace{-6mm}
  \caption{\textbf{Recursive Harness Distillation.} (1) A strong agent interacts with a frozen VLA and reflects on execution outcomes to identify useful intervention guidance. (2) This experience is distilled into a playbook that specifies when to intervene, how to adjust execution, and what to check afterward. The illustrated revision replaces a simple retry instruction with guidance to inspect the failed grasp, refocus on target–finger contact, and verify that the object follows the gripper. (3) A light agent consults the playbook to guide VLA execution through instruction, attention, and action interventions. Its execution experience returns to the strong agent for further playbook refinement, closing the recursive loop. }
  \label{fig:algorithm}
\end{figure}

In this paper, we propose \textbf{Recursive Harness Distillation (RHD)}, a framework that distills a capable agent’s experience into a reusable harness for a lighter agent (Figure~\ref{fig:algorithm}). The harness combines an interface for observing and intervening in VLA execution with a playbook that guides its use. A capable agent first interacts with the VLA and records effective strategies in the playbook. A lighter agent then executes tasks using this harness, and the capable agent reviews the resulting experience to revise the guidance. Through this repeated process, the harness captures both what works for the VLA and what the lighter agent needs to apply it.

We evaluate RHD on SimplerEnv Bridge and three real-world manipulation tasks. In real-world experiments, the harness improves success from 37.3\% to 64.0\%. On SimplerEnv Bridge, adding the light agent without a playbook yields only a modest improvement over the GR00T-only baseline, from 41.7\% to 43.8\%. With the playbook, the same light agent achieves 66.7\% success, outperforming the strong agent operating without a playbook. These benefits extend to the strong agent itself, which achieves 79.2\% success when equipped with the same playbook. Together, these results show that experience acquired through interaction with a VLA can be accumulated, refined, and reused across agents to improve robotic manipulation.

\section{Related Work}

\noindent\textbf{Vision-Language-Action Policies.}
VLA models have emerged as a promising approach to generalist robot policies, combining pretrained vision-language representations with learning from diverse robot demonstrations~\citep{brohan2023rt,kim2024openvla,black2024pi_0,bjorck2025gr00t}. Training on diverse manipulation data enables a shared policy to perform multiple tasks conditioned on language instructions~\citep{walke2023bridgedata,o2024open,team2024octo}. Their behavior depends on how instructions are grounded in visual observations and translated into motor commands. However, broad task coverage does not guarantee reliable execution under unfamiliar visual conditions, where existing models exhibit generalization limitations~\citep{zhou2025libero,liu2026vls}.
VLA models can struggle to translate learned manipulation capabilities into reliable behavior as scenes and execution states change. Our Recursive Harness Distillation enables an external agent to adapt its guidance to these conditions by accumulating and reusing experience from policy interventions.

\noindent\textbf{Harness.}
External control structures connect language-model reasoning to robot execution through interfaces for perception, action, and feedback~\citep{ahn2022can,huang2022inner,huang2023voxposer,yao2022react}. Code as Policies~\citep{liang2023code} uses language models to generate control programs that process perceptual outputs and compose robot APIs. More recent approaches incorporate learned policies as callable tools, enabling agents to decompose tasks, monitor progress, and intervene when execution fails~\citep{chen2026volo,galanti2026addressing}. These approaches provide mechanisms for guiding execution, but do not by themselves explain how effective intervention strategies can be accumulated and transferred across agents. Our RHD develops reusable guidance from policy interactions and refines it based on how subsequent agents apply it, with the goal of reducing repeated reasoning and trial and error.

\noindent\textbf{Knowledge Distillation.}
Knowledge distillation transfers a teacher's predictive behavior to a student through training targets~\citep{hinton2015distilling}, including teacher-generated rationales that supervise the student's reasoning~\citep{hsieh2023distilling}. Student feedback can also guide the transfer itself, informing teacher updates and revisions to the knowledge provided in subsequent rounds~\citep{liu2021learning,zhou2022bert,chen2025smaller}. These methods motivate adapting what is taught to what the student can use. A complementary line of work on LLM agents retains textual reflections and extracts reusable insights from task experience~\citep{shinn2023reflexion,zhao2024expel}, while further approaches organize and refine this knowledge through generation, reflection, and curation~\citep{zhang2026agentic}. RHD brings these ideas to VLA manipulation through a recursive teacher--playbook--student process. A strong agent initializes the playbook through policy interaction, then revises it using the light agent's subsequent executions. The object of refinement is external guidance for recognizing when and how to intervene on the VLA, grounded in the physical outcomes of the recipient agent's decisions.

\section{Method}
\label{sec:method}

\subsection{Problem Formulation}
\label{sec:formulation}
We consider manipulation tasks drawn from a distribution $\mathcal{D}$, each specifying a goal instruction $I$ and an initial scene. A pretrained VLA $\pi_\theta$ generates robot actions from observations and instructions. Its parameters $\theta$ remain fixed. An external agent observes execution and intervenes through an interface that exposes the instruction, internal feature attention, and action output. We denote the strong agent by $T$ and the lighter deployment agent by $S$. Their model parameters also remain fixed; knowledge acquired through execution is retained in a playbook $P$.

An episode has a finite horizon $H$. At step $t$, $h_t$ contains the goal, observations, previous interventions, and executed actions available up to that step. The underlying physical state need not be fully observed. The interface presents the relevant information from $h_t$ and makes a prescribed set of intervention operations available to the agent. Each operation includes an identity choice, so the agent can leave the corresponding part of the VLA unchanged. The playbook guides the agent's choice of operations and their arguments. Let $b_t$ denote the complete decision record at step $t$, including the chosen operations, the proposed VLA action, and the executed action. The agent and VLA together induce an execution policy $\Pi_{S,P}(b_t\mid h_t)$. Retaining this record makes the feedback available to subsequent decisions explicit. We seek a playbook that improves the task success of this composed policy.

\subsection{Intervening on VLA Execution}
\label{sec:interface}
We expose three intervention sites along the VLA's computation. An instruction operation transforms the language input, an attention operation modifies internal feature attention during inference, and an output operation transforms the generated action. At decision point $t$, let $o_t$ denote the current observation provided to the frozen VLA $\pi_\theta$, and let $I$ denote the original task instruction. The agent's instruction, attention, and output interventions are denoted by $u_t^I$, $u_t^F$, and $u_t^A$, respectively. For a candidate action sequence, these interventions are represented as

\begin{equation}
\widetilde I_t = f_I(I,u_t^I), \qquad
\widehat a_t = \pi_\theta(o_t,\widetilde I_t;u_t^F), \qquad
a_t = f_A(\widehat a_t,u_t^A).
\label{eq:interface}
\end{equation}

Here, $f_I$ maps the original instruction and the instruction intervention to the modified instruction $\widetilde I_t$. The frozen VLA produces the proposed action sequence $\widehat a_t$, and $f_A$ maps this proposal and the output intervention to the candidate action sequence $a_t$. The admissible transformations of $f_A$ are specified by the interface. The semicolon denotes an intervention on the VLA's forward computation: $u_t^F$ modifies selected attention computations over features while preserving the learned parameters $\theta$. A decision need not advance the environment: the agent may request and inspect alternative candidates at the same environment step. Upon approval, the executor applies only the selected prefix of the approved candidate, within the interface's execution limits, and returns a new observation. Thus, an agent decision does not necessarily correspond to executing an entire action chunk.

These intervention sites do not define a mandatory sequence of agent operations. The agent may select any admissible subset without first attempting the others. Instruction and attention interventions affect candidate generation, whereas output interventions modify candidate actions; this computational dependency does not impose an order in which the agent must try the intervention sites. Available operations can be combined within a decision; their identity settings recover the unmodified VLA. The choice of intervention site is part of the agent's decision. The playbook specifies the situations in which an operation is useful, how to apply it, and which subsequent observations indicate that it had the intended effect.

\subsection{Policy Optimization View: Improving the Recipient's Decisions}
\label{sec:optimization}
A playbook changes robot behavior through the agent that interprets it. We therefore view Recursive Harness Distillation as policy optimization over the policies induced by the deployment agent. Let $Y(\tau)\in\{0,1\}$ indicate whether trajectory $\tau$ achieves the original task goal within the horizon. The objective is
\begin{equation}
P^\star \in \arg\max_{P\in\mathcal{P}} J_S(P),
\qquad
J_S(P)=\mathbb{E}_{\tau\sim p_{\mathcal{D}}(\tau;\Pi_{S,P})}[Y(\tau)],
\label{eq:objective}
\end{equation}
where $\mathcal{P}$ denotes the set of admissible playbooks. Here, $p_{\mathcal{D}}(\tau;\Pi_{S,P})$ denotes the trajectory distribution induced by $\Pi_{S,P}$ on tasks and initial scenes drawn from $\mathcal{D}$.
The task distribution, VLA, intervention interface, and execution horizon are held fixed. The objective evaluates how the light agent uses the guidance, while the strong agent supplies experience and proposes revisions to it.

A playbook distilled from the strong agent's execution experience provides an initialization. Subsequent revisions use recipient rollouts, including failures, to refine when and how the guidance should be applied. The light agent can choose different interventions under the same guidance, and those decisions change the situations encountered later in the episode. The relevant experience distribution is thus induced by the recipient. This is closely related to the role of learner-induced distributions in sequential imitation learning~\citep{ross2011reduction}.

The dependence on the recipient's execution can be made explicit using the finite-horizon performance-difference identity~\citep{kakade2002approximately,schulman2015trust}. Let $k$ index playbook revisions. For the following identity, $t$ indexes agent--interface decisions, and $N(\tau)$ denotes their number before episode termination under the robot-step horizon $H$. A decision may generate a candidate without advancing the robot or execute a variable-length action prefix. For the current playbook $P_k$, let $V_k(h_t)$ be the probability of success when continuing with $\Pi_{S,P_k}$, and let $Q_k(h_t,b)$ be the success probability after the decision recorded in $b$ and then continuing with that policy. Define $\mathcal{A}_k(h_t,b)=Q_k(h_t,b)-V_k(h_t)$. For a candidate playbook $P'$, the change in success is

\begin{equation}
J_S(P')-J_S(P_k)
=
\mathbb{E}_{\tau\sim p_{\mathcal{D}}(\tau;\Pi_{S,P'})}
\left[
\sum_{t=0}^{N(\tau)-1}
\mathcal{A}_k(h_t,b_t)
\right].
\label{eq:improvement}
\end{equation}
The identity follows by telescoping the continuation values along the candidate policy's trajectory. It identifies the improvement target: a revision should induce better decisions at the histories the recipient actually encounters. Reproducing the teacher's decisions on its own histories alone does not evaluate this quantity. The values in Eq.~\ref{eq:improvement} characterize the objective; we evaluate proposed revisions through complete recipient rollouts and their task outcomes.

\subsection{Recursive Harness Distillation (RHD)}
\label{sec:distillation}
The strong agent first executes tasks through the intervention interface, producing experience $\mathcal{E}_T$. It extracts an initial playbook $P_0$ that relates observable situations to intervention choices and their observed consequences. The light agent then executes with $P_k$, producing experience $\mathcal{E}_{S,k}$. The strong agent reviews how the guidance was applied, which interventions were selected, and how the VLA and environment responded. It uses this evidence to propose revisions that address ambiguities, incorrect applicability conditions, and ineffective intervention choices.

We formulate the update as a search over playbooks proposed by the teacher. Let $\mathcal{G}_T(P_k,\mathcal{E}_T,\mathcal{E}_{S,0:k})$ denote a finite set of proposed revisions based on the available teacher experience $\mathcal E_T$ and accumulated recipient experience $\mathcal E_{S,0:k}$, where $\mathcal E_{S,0:k}$ denotes experience collected from revisions $0$ through $k$. Proposed revisions may be generated and evaluated sequentially.
Let $\mathcal C_k$ collect the current playbook and the revisions proposed during the search. The set is constructed incrementally as proposals are generated:

\begin{equation}
\mathcal{C}_k=\{P_k\}\cup
\mathcal{G}_T(P_k,\mathcal{E}_T,\mathcal{E}_{S,0:k}),
\qquad \mathcal{C}_k\subseteq\mathcal{P}.
\label{eq:proposals}
\end{equation}

Each proposed revision is fixed before evaluation with the light agent on a development batch of $n$ task instances. Within a development round, the same instances are reused across evaluated revisions. Denoting these rollouts by $\tau_i^{S,P}$, we measure empirical success. When a candidate meets the target, the accepted playbook satisfies:

\begin{equation}
P_{k+1}\in
\left\{
P\in\mathcal C_k:
\widehat J_{S,k}(P)\geq\eta
\right\},
\qquad
\widehat J_{S,k}(P)
=
\frac{1}{n}\sum_{i=1}^{n}Y(\tau_i^{S,P}).
\label{eq:selection}
\end{equation}

Here, $\eta$ is the teacher's empirical success rate on the same development instances. Candidates are proposed and evaluated sequentially, and the first candidate satisfying Eq.~(\ref{eq:selection}) is accepted. If no candidate meets the target, execution feedback guides further proposals. The acceptance rule does not guarantee termination; until a candidate meets the target, no update is accepted under this rule. Subsequent rounds collect new experience under the revised playbook. The resulting recursion couples knowledge transfer to its actual use: teacher experience starts the process, and recipient experience continually informs what the teacher should convey next.

\section{Experiments}
\label{sec:experiments}
We evaluate whether experience distilled into a playbook improves the use of a frozen VLA, and whether these gains extend from a light agent to a more capable agent and to physical robot execution.

\subsection{Experimental Setup}
\begin{figure}[t]
    \centering
    \includegraphics[width=\linewidth]{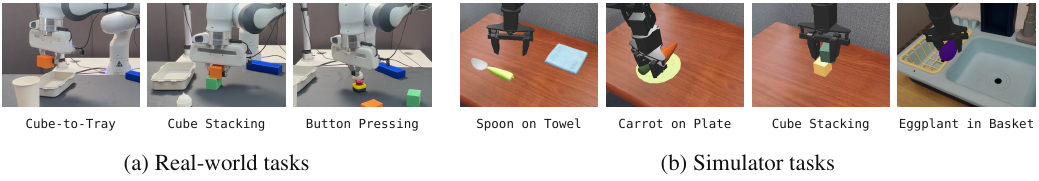}
    \vspace{-7mm}
    \caption{\textbf{Experimental setups.}
    (a) Three real-world tasks: Cube-to-Tray, Cube Stacking, and Button Pressing.
    (b) Four SimplerEnv Bridge tasks: Spoon-on-Towel, Carrot-on-Plate, Cube Stacking, and Eggplant-in-Basket.}
    \label{fig:experimental_setups}
\end{figure}

\noindent\textbf{Simulation Setup.}
We evaluate on four WidowX manipulation tasks in the Bridge setting of SimplerEnv~\citep{li2024evaluating}: spoon-on-towel, carrot-on-plate, cube stacking, and eggplant-in-basket (Figure~\ref{fig:experimental_setups}b). We use the GR00T-N1.7-SimplerEnv-Bridge checkpoint and keep its parameters fixed throughout playbook development and evaluation. The policy receives a $256\times256$ RGB image, robot proprioception, and a language instruction. We reserve 12 initial configurations for development and 12 for evaluation, yielding 48 instances per split.

\noindent\textbf{Real-World Setup.}
We use a 7-DoF Franka Panda with wrist, front, and right-side RGB cameras for three tasks (Figure~\ref{fig:experimental_setups}a): Cube-to-Tray, which places an orange cube in a tray; Cube Stacking, which places the orange cube on a green cube; and Button Pressing, which presses a red emergency stop button. We collect 50 human-teleoperated demonstrations per task, randomizing object positions across demonstrations, and fine-tune $\pi_{0.5}$~\citep{Intelligence202505AV} from its base checkpoint. The fine-tuned policy is then fixed, and the harness uses the same underlying policy execution settings as the policy-only baseline.
At evaluation time, object positions are randomized before every trial within a $50\mathrm{cm}\times50\mathrm{cm}$ workspace, while remaining visible from the wrist camera. This requires the policy to handle variation in object placement across trials. We report success over 25 trials per task, for 75 trials in total. For real-world evaluation, Astra adapts the GR00T-derived playbook to the physical tasks and fine-tuned \(\pi_{0.5}\) policy, and Luna uses the adapted guidance.

\noindent\textbf{Agents.}
We use GPT-6 Astra as the strong agent and GPT-5.6 Luna as the light agent. Astra collects interaction experience with GR00T and revises the playbook using recorded observations, interventions, and outcomes from Luna’s execution. Agents can replace the policy instruction with text of up to 512 characters, apply an attention-logit bias of strength 0–4 to visual tokens within a selected image region, and modify end-effector motion and gripper commands for up to five control steps per intervention. The intervention parameters are selected by the agent from the current observations. Later development rounds combine eight new teacher cases, two per task, with eight previously encountered cases for recipient evaluation. Revisions compare teacher and recipient execution on the same cases to clarify intervention conditions and outcome checks. Each revised playbook is saved as a separate version before subsequent execution. The initial checkpoint is distilled from Astra’s first eight development rollouts with GR00T. Luna then executes development tasks using this playbook, and Astra revises the guidance using Luna’s execution feedback. Subsequent checkpoints incorporate additional interaction experience and recipient feedback.

\noindent\textbf{Evaluation Protocol.}
Playbook construction and refinement use development instances, while evaluation uses initial configurations excluded from that process. Each evaluated playbook is frozen across all evaluation episodes. Maximum episode lengths are 120 steps for spoon-on-towel, carrot-on-plate, and cube stacking, and 240 steps for eggplant-in-basket. Comparisons use the same evaluation configurations, reset seeds, and policy checkpoint. Agents without a playbook receive interface documentation. Simulator object states and success signals are excluded from agent inputs; success is determined by the environment against the original task goal, even when the agent modifies the policy instruction.

\subsection{Main Results}
\noindent\textbf{SimplerEnv Bridge.}
\begin{figure}[t]
    \centering
    \includegraphics[width=\linewidth]{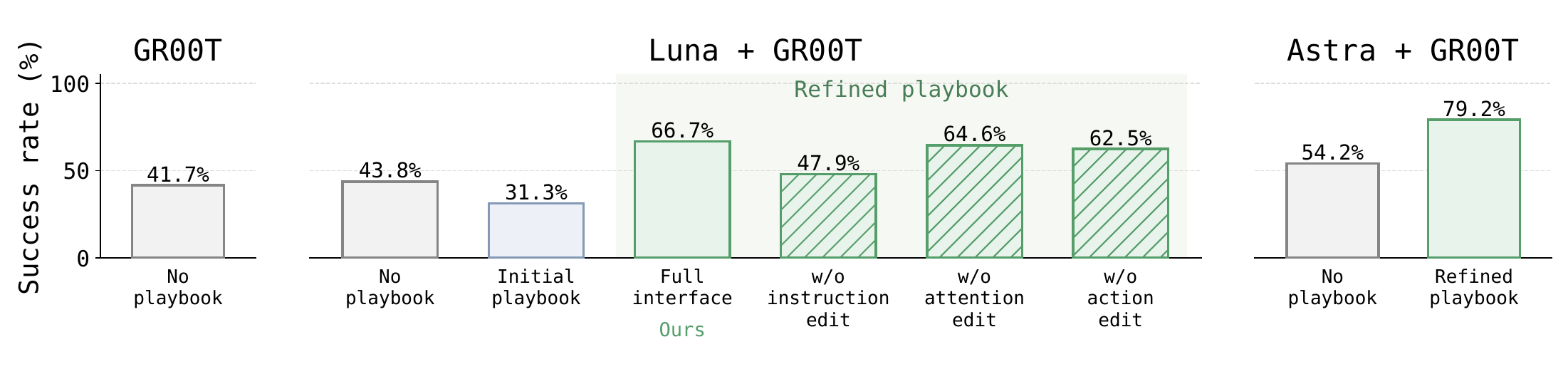}
    \vspace{-7mm}
    \caption{
       \textbf{Overall execution success and intervention ablations.}
We compare agents without a playbook and with initial or
refined playbooks, and examine the contribution of each
intervention site for Luna + GR00T.
Each ablation disables one site while retaining the other
two and keeping the refined playbook fixed.
The results suggest that the gains extend beyond direct
action correction, with playbook-guided instruction editing
making the largest contribution among the tested sites.
    }
    \label{fig:bridge-overall-ablation}
\end{figure}

The refined playbook substantially improves execution with the same agent, policy, and interface (Figure~\ref{fig:bridge-overall-ablation}). GR00T alone achieves 41.7\% success, and adding Luna without a playbook raises this to 43.8\%. With the refined playbook, Luna reaches 66.7\%, a gain of 22.9 percentage points over its unguided execution. The same playbook also benefits Astra, which reaches 79.2\%. Experience accumulated through policy interaction therefore produces guidance that is useful across agents with different capabilities.

\noindent\textbf{Real-World Manipulation.}
\begin{table}[t]
\centering
\setlength{\belowcaptionskip}{6pt}
\caption{\textbf{Real-world manipulation success rates.}
We compare the frozen policy with Luna-guided execution,
with and without the refined playbook, over 25 trials per task.
Task entries report successful trials; Overall reports
the aggregate success rate (\%).}
\label{tab:real_world}
\small
\renewcommand{\arraystretch}{1.15}
\begin{tabular*}{\linewidth}{
  @{\extracolsep{\fill}} llcccc @{}
}
\toprule
Method & Playbook &
\shortstack{Cube-to-Tray} &
\shortstack{Cube Stacking} &
\shortstack{Button Pressing} &
Overall (\%)\\
\midrule
$\pi_{0.5}$ & None
& 15/25 & 3/25 & 10/25 & 37.3 \\
Luna + $\pi_{0.5}$ & None
& 0/25 & 0/25 & 0/25 & 0 \\
\textcolor{oursgreen}{\textbf{Luna + $\pi_{0.5}$}} & \textcolor{oursgreen}{\textbf{Refined (Ours)}}
& 18/25 & 13/25 & 17/25 & \textbf{64.0}\\
\bottomrule
\end{tabular*}
\end{table}

We evaluate the harness on a Franka Panda using $\pi_{0.5}$ for Cube-to-Tray, Cube Stacking, and Button Pressing (Table~\ref{tab:real_world}). Across the three tasks, the playbook improves Luna's overall success to 64.0\%, compared with 37.3\% for $\pi_{0.5}$ alone and 0\% for Luna with the policy but no playbook. These results show that access to intervention tools alone does not ensure effective control of the physical robot. Without a playbook, Luna struggled to compensate for recurring spatial offsets between the intended interaction location and the robot's executed motion. Changes to instructions or attention did not reliably resolve these errors in the task-specific fine-tuned policy. The real-world playbook was adapted by Astra from the GR00T-derived playbook to the current physical tasks, providing Luna with task-specific guidance for intervening during execution.

\subsection{Recursive Harness Distillation}
\label{sec:teacher_model}
\begin{table*}[t]
\makeatletter
\renewcommand{\fnum@table}{\small\tablename~\thetable}
\renewcommand{\fnum@figure}{\small\figurename~\thefigure}
\makeatother
\centering
\hyphenpenalty=10000
\exhyphenpenalty=10000

\begin{minipage}[t]{0.40\textwidth}
\vspace{0pt}
\centering
\hyphenpenalty=10000
\exhyphenpenalty=10000
\setlength{\abovecaptionskip}{0pt}

\caption{\small\raggedright
\textbf{Contribution of the teacher model.}
Success on 48 held-out SimplerEnv Bridge instances
using refined playbooks.
All recipients operate frozen GR00T;
teacher and recipient use separate contexts.
}
\label{tab:teacher_model}

\vspace{6pt}
\begingroup
\small
\setlength{\tabcolsep}{3pt}
\renewcommand{\arraystretch}{1.15}

\begin{tabular*}{\linewidth}
{@{\extracolsep{\fill}}lcc@{}}
\toprule
Teacher & Recipient & Overall (\%) \\
\midrule
Luna & Luna & 22.9 \\
\textcolor{oursgreen}{\textbf{Astra}}
& \textcolor{oursgreen}{\textbf{Luna}} & \textbf{66.7} \\
\bottomrule
\end{tabular*}

\endgroup
\end{minipage}
\hfill
\begin{minipage}[t]{0.57\textwidth}
\vspace{0pt}
\centering
\hyphenpenalty=10000
\exhyphenpenalty=10000

\includegraphics[width=\linewidth]{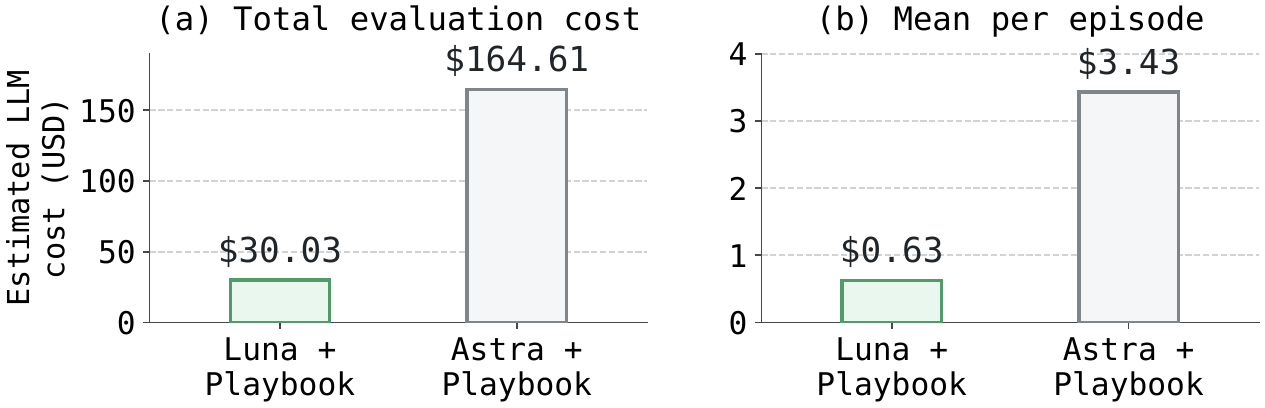}\par
\setlength{\abovecaptionskip}{2pt}

\makeatletter
\def\@captype{figure}
\makeatother
\caption{\small\raggedright
\textbf{Agent inference cost.}
Total and per-episode API cost estimates for
Luna and Astra operating GR00T with the refined playbook on the same test set.
}
\label{fig:deployment_cost}

\end{minipage}
\end{table*}

\begin{figure*}[t]
    \centering
    \includegraphics[width=\textwidth]{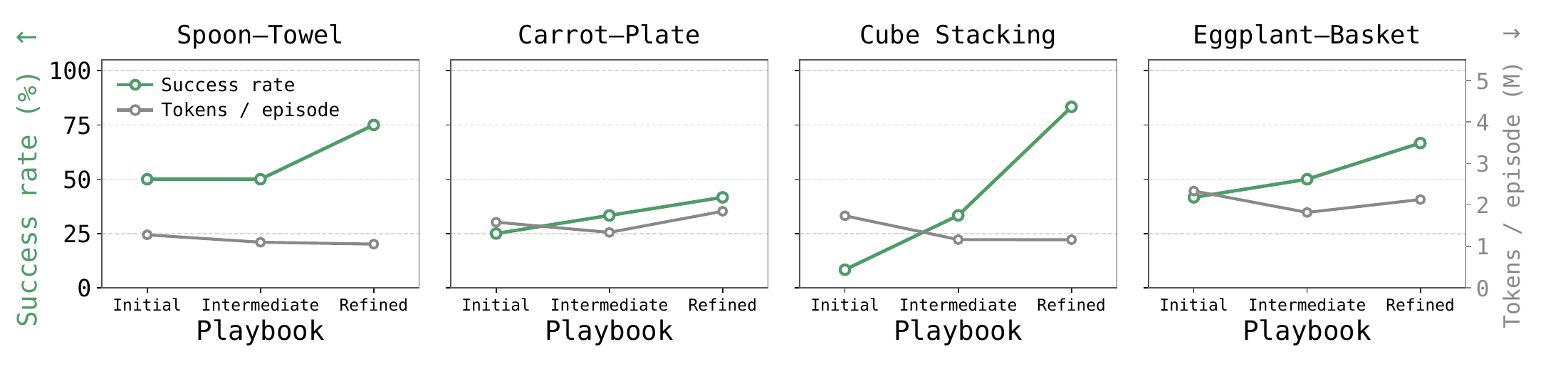}
    \vspace{-7mm}
    \caption{\textbf{Task success and deployment token use across playbook revisions.}
    Success rate (left axis) and mean agent token usage per episode (right axis) are evaluated on the same 12 held-out instances per task. Token usage is averaged over all episodes, including failures.}
    \label{fig:playbook_success_tokens}
\end{figure*}

\noindent\textbf{Effect of Iterative Refinement.} The initial playbook reduces Luna’s success from 43.8\% without a playbook to 31.3\%, whereas the refined playbook raises it to 66.7\%. Figure~\ref{fig:playbook_success_tokens} tracks success rates
and mean tokens per episode across the initial, intermediate, and refined playbooks, showing that improvements emerge at different stages across tasks.
Refinement improves success across all four tasks while reducing mean tokens per episode in most tasks relative to the initial playbook. Overall, playbook refinement improves success without increasing average token usage per episode.

\noindent\textbf{Role of a More Capable Teacher.} To examine whether a Luna teacher can obtain the same gains through recursive refinement, we compare Luna and Astra as teachers while keeping Luna as the recipient operating the same frozen GR00T on SimplerEnv Bridge tasks (Table~\ref{tab:teacher_model}). In the Luna-to-Luna setting, the teacher and recipient operate in separate contexts. Both settings use the same development split, interface, playbook length limit, and limits on development rollouts and revision attempts. The Luna-derived refined playbook achieves only 22.9\% success, compared with 66.7\% for the Astra-derived counterpart, and falls below Luna without a playbook (43.8\%). These results show that having the light agent construct and recursively refine its own playbook is insufficient to obtain the gains achieved with a strong teacher. This comparison supports the teacher--recipient hierarchy in RHD: a more capable teacher develops and refines guidance that the lighter recipient can reuse.

\subsection{Contribution of Intervention Sites}
\label{sec:intervention_ablation}
We evaluate the contribution of instruction editing, attention guidance, and action correction by disabling one intervention site at a time while retaining the other two and keeping the refined playbook fixed (Figure~\ref{fig:bridge-overall-ablation}). Removing instruction editing causes the largest performance drop among the tested sites. The results show that playbook-guided use of the policy instruction contributes to the gains alongside direct action correction.

\section{Analysis}
\label{sec:analysis}
We examine what the playbook captures about the policy, how this guidance changes execution, and the inference cost of deploying the recipient.

\subsection{Understanding Policy Behavior through the Playbook}
\label{sec:example}
\begin{figure}[t]
    \centering
    \includegraphics[width=\linewidth]{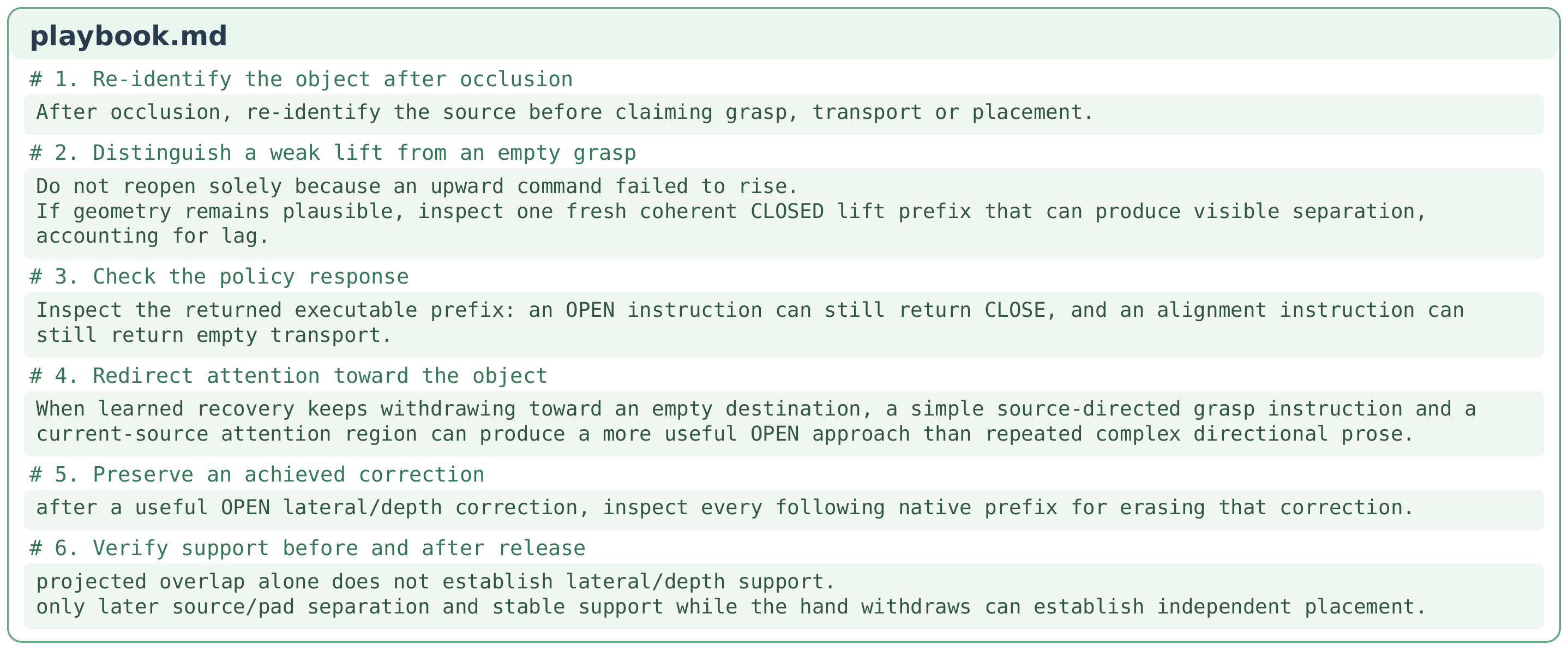}
    \vspace{-7mm}
    \caption{\textbf{Operational insights from the playbook.} Selected excerpts address object identity after occlusion, weak versus empty lifts, instruction--action mismatches, attention-based recovery, preservation of corrections, and evidence for placement.}
    \label{fig:playbook_examples}
\end{figure}

Figure~\ref{fig:playbook_examples} presents six examples from the playbook developed through interactions with GR00T on SimplerEnv Bridge tasks, illustrating how execution experience becomes practical intervention guidance.

\textbf{Interpreting Observations.} During development, Luna mistook a robot component for the spoon after occlusion, motivating explicit object re-identification using appearance and scene landmarks (entry 1). In eggplant manipulation, a weak initial lift was followed by a closed-gripper lift that confirmed possession. The playbook therefore recommends an additional bounded lift when the grasp remains plausible, rather than reopening solely because upward motion is small (entry 2).

\textbf{Selecting Interventions.} An OPEN instruction can still produce closing actions, so the playbook requires inspecting policy outputs and applying direct action corrections when needed (entry 3). When the policy moves toward the destination without the object, it recommends a simple grasp instruction with attention on the object's current location. In one cube-recovery sequence, this combination changed an upward proposal into an open-gripper descent; the proposed motion and gripper state still require inspection (entry 4).

\textbf{Verifying Physical Outcomes.} In a cube-stacking episode, executing the native prefix, the initial portion of a VLA-predicted action sequence, reversed a useful position correction. The playbook therefore requires checking that the corrected grasp geometry is preserved before closing the gripper (entry 5). Later, releasing the cube without confirmed support caused it to fall in front of the target. The playbook therefore requires checking object–target geometry before release and, when execution continues, stable support as the gripper withdraws (entry 6). These checks turn interaction experience into explicit conditions for selecting and verifying interventions.

\subsection{How Guidance Changes Execution}
\label{sec:qualitative}
\begin{figure*}[t]
    \centering
    \includegraphics[width=\textwidth]{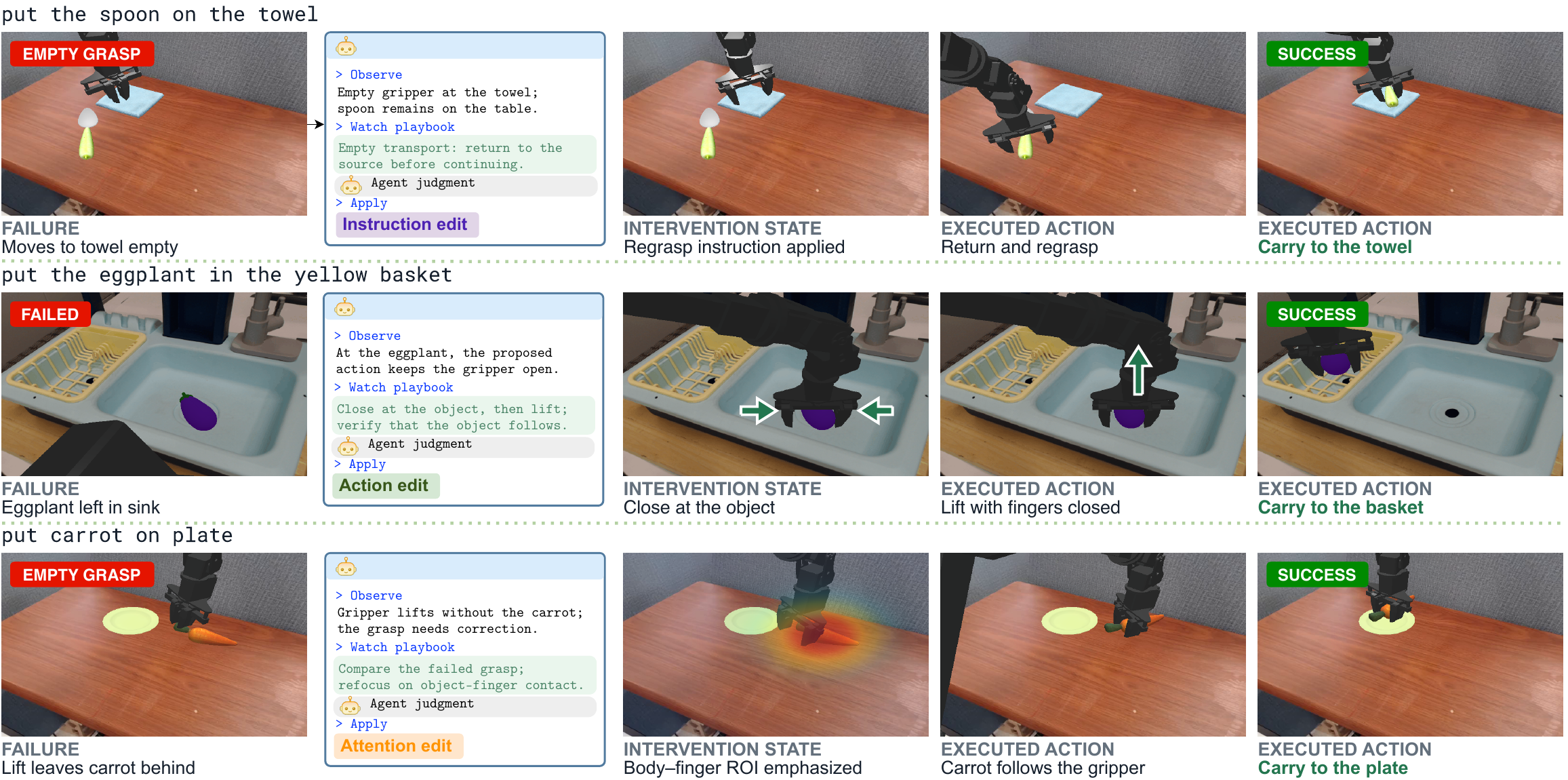}
    \vspace{-6mm}
    \caption{\textbf{Playbook-guided recovery in three Bridge tasks.}
    Each row connects an observed failure, a relevant playbook rule, an intervention, and subsequent execution. Multiple interventions can contribute to a recovery.}
    \label{fig:qualitative}
\end{figure*}

The examples show how an agent can return a stalled execution to a state from which GR00T can continue the task (Figure~\ref{fig:qualitative}). For the spoon, the instruction edit directs the robot back to the source to regrasp before returning to the towel. For the eggplant, correcting the gripper action allows the robot to lift and carry the object. For the carrot, the spatial intervention focuses on object--finger contact before transport resumes. These cases illustrate how guidance about a local failure can support completion of the original task through the existing policy.

\subsection{Deployment Cost}
\label{sec:cost}
Agent inference cost is an important consideration for repeated robot deployment. We compare Luna and Astra operating frozen GR00T on the same held-out Bridge test set using the refined playbook. Costs are calculated from recorded token usage and model-specific API rates (Figure~\ref{fig:deployment_cost}). Per million tokens, Luna is priced at \$0.20 for standard input, \$0.25 for cache writes, \$0.02 for cache reads, and \$1.20 for output; Astra is priced at \$10 for uncached input, \$1 for cached input, and \$50 for output. We normalize the total cost by the number of evaluation episodes to obtain the mean cost per episode. Over the full test set, Luna incurs an estimated inference cost of \$30.03, compared with \$164.61 for Astra. The corresponding mean costs are \$0.63 and \$3.43 per episode, respectively, making Astra approximately 5.5 times more expensive. This cost gap motivates deploying a light agent that reuses guidance distilled from a strong agent through the playbook.

\section{Conclusion}
We presented Recursive Harness Distillation, a framework for accumulating a strong agent's experience with a VLA as reusable guidance for a light agent. The strong agent distills its interaction experience into a playbook and recursively refines it using the light agent's execution feedback, adapting the guidance to how the recipient applies it without updating model parameters. Experiments on SimplerEnv Bridge and three real-world manipulation tasks show that the resulting playbook improves the light agent's ability to operate the deployed policy in both simulation and physical environments. On Bridge, the light agent with the playbook outperforms the strong agent without one, while the same playbook also improves the strong agent's performance. These results demonstrate the feasibility of harness distillation for robotics: experience with a deployed policy can be accumulated, refined through execution, and reused across agents to improve manipulation.

\bibliography{references}
\bibliographystyle{iclr2027_conference}

\end{document}